\documentclass[runningheads]{llncs}
\usepackage{graphicx}
\usepackage{amsmath}
\usepackage{amsfonts}
\usepackage{multirow}
\usepackage{algorithmic}
\usepackage{array}
\usepackage{fixltx2e}
\usepackage{url}
\usepackage{hyperref}
\usepackage{color}
\usepackage{cite}
\begin{document}
\title{An Effective Single-Image Super-Resolution Model Using Squeeze-and-Excitation Networks}
\titlerunning{SrSENet}
%
\author{Kangfu Mei\inst{1}\orcidID{0000-0001-8949-9597} \and
Aiwen Jiang \thanks{Corresponding author} \inst{1}\orcidID{0000-0002-5979-7590} \and \\
Juncheng Li\inst{2}\orcidID{0000-0001-7314-6754} \and
Jihua Ye\inst{1} \and
Mingwen Wang\inst{1}}
\authorrunning{Kangfu Mei, Aiwen Jiang, et al.}
%
\institute{School of Computer Information Engineering, Jiangxi Normal University\\
\email{\{meikangfu, jiangaiwen, jhye, mwwang\}@jxnu.edu.cn}\\
\and
Department of Computer Science \& Technology, East China Normal University\\
\email{51164500049@stu.ecnu.edu.cn}}
\maketitle              
\begin{abstract}
    Recent works on single-image super-resolution are concentrated on improving performance through enhancing spatial encoding between convolutional layers.
    In this paper, we focus on modeling the correlations between channels of convolutional features.
    We present an effective deep residual network based on squeeze-and-excitation blocks (SEBlock) to reconstruct high-resolution (HR) image from low-resolution (LR) image.
    SEBlock is used to adaptively recalibrate channel-wise feature mappings.
    Further, short connections between each SEBlock are used to remedy information loss.
    Extensive experiments show that our model can achieve the state-of-the-art performance and get finer texture details.
\end{abstract}
\section{Introduction}
Single-image super-resolution(SISR) is a popular computer vision problem, which aims to reconstruct a high-resolution (HR) image from a low-resolution(LR) image.
However, SISR is still considered as an ill-posed inverse problem due to high-level information loss during image downsampling.
To solve this problem, many algorithms have been proposed.

Early methods~\cite{timofte2014a+, Yang2013, yang2010image, yang2008image, schulter2015fast}, besides bicubic and bilinear interpolation, learned the mapping from LR to HR pairs directly by sacrificing certain accuracy or speed for improvements. Super-Resolution Convolutional Neural Network (SRCNN) proposed by Dong et al.~\cite{dong2014learning} was the first successful model that adopted CNN structure to solve SISR problem and obtained great performance improvement.
In SRCNN, convolutional neural network was used to learn non-linear mapping from each LR vector to a set of HR vector.
Due to the outstanding performance of SRCNN, several deeper and more complicated models has been proposed to follow it, such as VDSR proposed by Kim et al.~\cite{kim2016accurate}. Though VDSR achieved excellence performance, its speed remained slow speed as it use a very deep residual convolutional network and an upscale image preprocess.

To avoid the complexities of feature extraction network and upscale preprocess, Shi et al.~\cite{shi2016real} replaced upscale preprocess with sub-pixel convolution layers. The sub-pixel layers could produce HR image from feature maps directly with a set of up-scaling filters. This architecture greatly improved the speed of networks. Therefore, following the strategy of up-sampling layer, Ledig et al.~\cite{ledig2016photo} further proposed a SRResNet with a very deep ResNet~\cite{he2016deep} architecture. Lai et al.~\cite{LapSRN} proposed the LapSRN, which use learned kernel as up-sampling unit to direct produced SR images.

In spite of great success achieved in the above architectures, the main issue that how to model mapping from LR to HR images better in a fast and flexible way remained unsolved.
In this paper, we have proposed a Super-Resolution Squeeze-and-Excitation Network (SrSENet) for SISR. The concept of SEBlock~\cite{hu2017squeeze} is employed to better modeling interdependencies between channels. Short connections from input to each SEBlock are used to remedy information lost. And different deconvolution layers are used for different scales under the same feature extraction architecture. The proposed method is evaluated on some popular publicly available benchmarks. Extensive experiments show that our proposed model can achieves competitive accuracy in a more accurate and flexible way. It can greatly reduce model’s complexity by using less layers and allow designing more flexible applications.

The contributions of this paper are two folds:
\begin{itemize}
    \par \item We have introduced an effective super-resolution network with SEBlock.
    It performs dynamic channel-wise feature recalibration to provide a new powerful architecture to improve the representational ability of information extraction part from low-resolution images.
    \par \item We have set up a new state-of-the-art super-resolution method with fast running speed and accurate result in the measurement of PSNR and SSIM without increasing the complexity of the network, especially in case of large upscale rate.
\end{itemize}

\section{Related Work}
\subsection{Single-Image Super-Resolution}
In this section, we are mainly concentrated on reviewing mainstream deep learning based single-image super-resolution methods. Typically, a SISR network could be approximately divided into two parts. The first part could be seen as a feature extraction block, which is composed of many stacked convolutional layers. The second part records up-scaling information from LR images to HR images. Recent works are concentrated on improving the first part by changing the way of skip connections between inputs of each layer. In other words, they focus on changing the proportion of information captured by initial layers.

\begin{figure}[ht]
    \setlength{\abovecaptionskip}{0.cm}
    \setlength{\belowcaptionskip}{-0.cm}
    \begin{center}
    \includegraphics[width=4in]{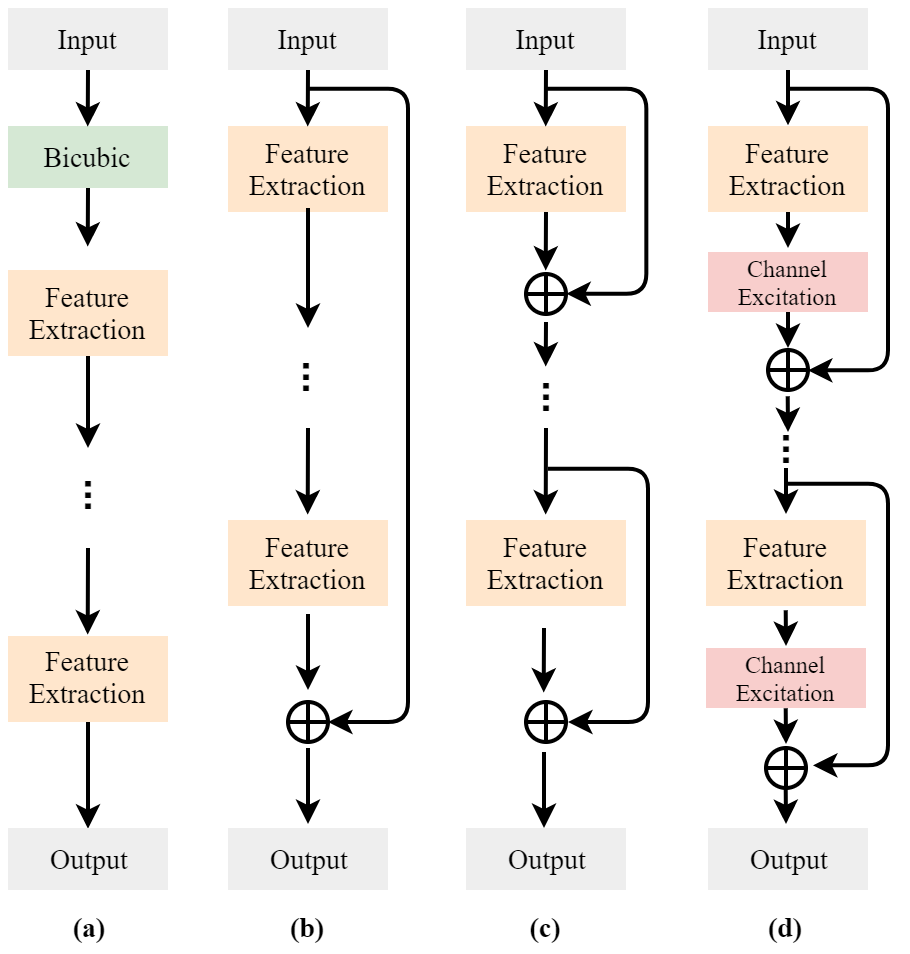}
    \end{center}
    \caption{Comparisons on network architectures of four typical deep learning based SISR categories.}
    \label{compare}
\end{figure}

We group mainstream deep learning based SISR models into four categories, as shown in Figuree~\ref{compare}. The (a) category contains feature extraction, such as network in \cite{dong2014learning}. The (b) category like \cite{kim2016accurate} introduces short connection as residual-learning. The (c) category like accepts input in each feature extraction layer. Our proposed model could be categorized into the last category (d). The difference from the other three categories is that each extraction layer block receives input before channel-wise modeling. In this way, network could better learn mapping between LR-HR images.

\subsection{Squeeze-and-Excitation Channel}

Different from works on enhancing spatial encoding, SENet~\cite{hu2017squeeze} was proposed to fully capture channel-wise dependencies through adaptive recalibration. The SENet was separated into two steps, squeeze and excitation, to explicitly model channel interdependencies.

\begin{figure*}
    \setlength{\abovecaptionskip}{0.cm}
    \setlength{\belowcaptionskip}{-0.cm}
    \begin{center}
    \includegraphics[width=4.8in]{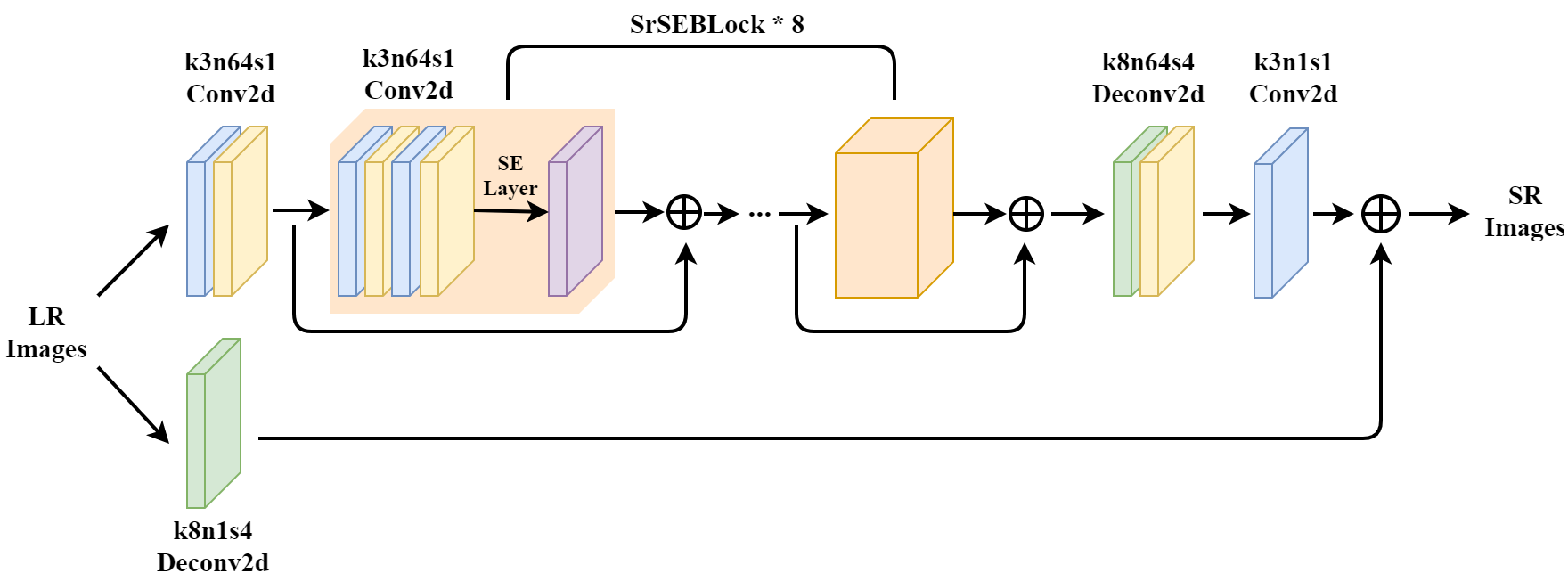}
    \end{center}
    \caption{Our proposed Network architectures of SrSENet in upscale of 4x. Blue blocks represent a Convolutional layer. Yellow blocks represent a LeakRelu layer. Green Blocks represent a Transposed Convolutional layer.}
    \label{SrSENet}
\end{figure*}

After initial images were input into the first convolution layer, the output feature $ U \in \mathbb{R}^{W \times H \times C}$ was passed to a SEBlock to do squeeze and excitation operator. The squeeze operator was used to embed information from global receptive field into a channel descriptor in each layer. Then a sigmoid activation function and FC layer were later used to gain nonlinear interaction between each layers. The squeeze operator produced a sequence S in $1 \times 1 \times C$ which represented the correlations of each layer.
The excitation operator later was employed to perform feature recalibration through reweighting the original feature mappings
$$ \tilde{U} = F_{scale}(U, S) = u_c \times s_c , $$
where $u_c$ refers to the parameters of the $c$-th filter and $s_c$ denotes the element of $c$-th channel descriptor. This architecture can help feature extaction parts better caputre the information from input to output. In our work, we combined SEBlock with ResNet for feature extraction.

\subsection{Transposed Convolutional Layer}
In order to obtain super-resolution images, a simple idea is to upscale original image first, then final HR image is directly generated from the resulted scaled image. It is not difficult to find that this kind of strategy wastes much time on preprocessing without any obvious advantage.

Shi et al.~\cite{shi2016real} first proposed to use sub-pixel convolution layer to produce HR images directly. It upscale a LR image by periodic shuffling the elements of a $W \times H \times C \cdot r^2$ tensor to a tensor of shape $rH \times rW \times rC$. However, it didn't make full use of the correspondence information from LR to HR.
LapSRN~\cite{LapSRN} was proposed by to use a multiple transposed convolutional layer to deal with different upscale rate in a progressive way. Without any preprocessing step like upscale, LapSRN achieved more accurate information between LR and HR in a fast way.

Following previous works, we use transposed convolutional layer with different parameters for different upscale rate, which can keep network simple and improve the power of networks to record reconstruction information.

\section{Proposed Method}
The proposed method aims to extract information from the LR image $I_{L}$ and
learn mapping function $F$ from $I_{L}$ feature maps to HR images $I_{H}$. We describe $I_{L}$ with $C$ channels in size of $W \times H$. With upscale rate $r$, $I_{H}$ is in size of $rW \times rH$. Our ultimate goal is to minimize
the loss between the reconstructed images and the corresponding ground truth
HR images. In the following, we will describe the details of the proposed method.

\subsection{Network Architecture}
Our proposed method is inspired from SRResNet~\cite{ledig2016photo} and LapSRN~\cite{LapSRN}. Following LapSRN, our model contains two parts: residual learning stage and image reconstruction stage, as shown in Figure~\ref{SrSENet}.

Unlike SRResNet and LapSRN, in the residual learning stage, we introduced SrSEBlock to extract features from LR images. The SrSEBlock structure integrates ResNet and SENet, which can better capture information from inputs and better modeling interdependencies between channels.

As VDSR~\cite{kim2016accurate} suggested, in the SR ill-posed problem, surrounding pixels were useful to correctly infer center pixel. With larger receptive field a SR model has, it could use more contextual information from LR to better learn correspondences from LR to HR. In our proposed network, the filters of SrSEBlock is in size of $3\times3\times64$. Therefore, in case of depth D layer, its receptive field could be seen as $(2\times D+1)\times(2\times D+1)$ in the original image space. The bigger receptive field means our network can use more context to reconstruct images.

As we know, with the increase of network depth, gradient disappearance or explosion will occur during training and the high-frequency information will also disappear. So, we introduce a short connection between SrSEBlocks which can receive input information before channel-wise modeling.

In the proposed network, we employ 8 SrSEBlocks to generate a feature mapping, and then we employ a transposed layer to transform the resulted mapping directly into a residual image by applying a deconvolutional layer. On different upscale rates, we don't increase the number of deconvolution layers, just directly change parameters such as the kernel size, stride and padding steps, to obtain corresponding residual image.
In image reconstruction stage, the up-sampled LR image feature mappings and the learned residual feature mappings are added together to reconstruct HR image. By using residual image learning, network converges efficiently. The final feature mapping is output directly as the SR image.

\begin{figure}
    \setlength{\abovecaptionskip}{0.cm}
    \setlength{\belowcaptionskip}{-0.cm}
    \begin{center}
    \includegraphics[width=2.4in]{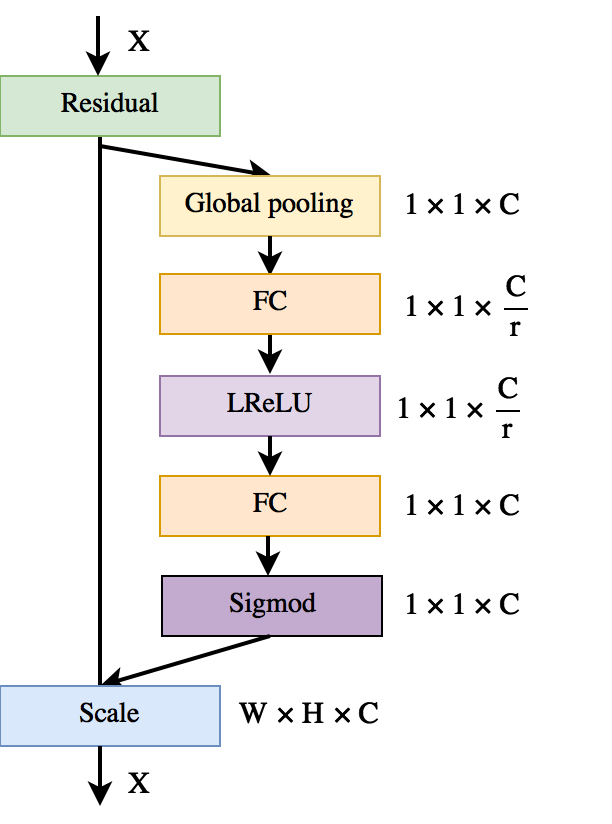}
    \end{center}
    \caption{The architecture of SELayer.}
    \label{SELayer}
\end{figure}
\subsection{Channels Excitation in SrSEBlock}
  Different from recent work that focus on enhancing spatial encoding, we use SrSEBlock to model correlations between channels. In this section, we will describe how the SrSEBlock work in our network.

  In details, feature maps are input into a SELayer as Figure~\ref{SELayer} shows. The corresponding excitations to each channel are output to scale original feature map. Taking a feature maps $U$ in size of $W\times H\times C$ as input, we first do a global average pooling to generate channel-wise statistics $z$ in size of $1 \times 1 \times C$., as show in below
  $$ z_{c} = \frac{1}{W\times H}\sum_{i=1}^W \sum_{j=1}^H u_{c}(i,j) .$$

  In order to learn nonlinear interaction between each channels, we use two FC layers with non-linear activations to form a bottleneck, as done in He et al.~\cite{he2016deep}. This architecture could limit model complexity and benefit for generalization. The reduction ratio r at 16 is accepted to do dimensionality reduction. The final output s of SELayer is use to scale corresponding channels of residual feature mappings.

  In this way, noise information in previous feature mappings could be reduced. And channels that contain useful information will be highly activated, helping to boost feature's discriminative abilities. In the later ablation experiment, we will show its effectiveness.

  \section{Experiments}
  In our experiment settings, given a set of HR images $ \{Y_{i}\} $ and the corresponding down-sampled LR images $ \{X_{i}\} $ through bicubic, our goal is to minimize the Charbonnier Penalty Function~\cite{bruhn2005lucas} defined as below, which is a differentiable variant of $L_{1}$ norm
  $$ \rho(z) = \sqrt{z^{2}+\varepsilon^{2}} . $$

  The loss is minimized using stochastic gradient descent
  with the standard backpropagation. We solve:
  $$ {G}^{*} = arg\,\min\limits_{G}\, \frac{1}{n} \sum_{i=1}^{n} \rho(Y_{i} - G(X_{i})) , $$
  where $G$ represents our SR image networks.

  \subsection{Datasets for Training and Testing}

  Different from previous work, we use DIV2K~\cite{Agustsson_2017_CVPR_Workshops} to train our model for more realistic modeling. DIV2K is a newly distributed high quality image dataset for image super resolution. Its training data has 800 high definition, high resolution images. In our experiments, we find different image processing framework will produce different bicubic downscale results. So for fair comparison, we all use the bicubic downsampling algorithm in Matlab image processing tool to generate LR-HR image pairs for our network training. For each pair, we crop HR sub image in $96 \times 96$ size and downscale it to LR images by different downscale factors. We export the pairs as MAT variable in HDF5 type.
  \subsection{Experiment Setup}
  We compare our proposed SrSENet with several state-of-the-art methods such as SRCNN~\cite{dong2014learning}, FSRCNN~\cite{dong2016accelerating}, SelfExSR~\cite{huang2015single}, VDSR~\cite{kim2016accurate}, DRCN~\cite{kim2016accurate} and LapSRN~\cite{LapSRN} on five common used benchmark datastes Set5, Set14\cite{zeyde2010single}, BSDS100\cite{martin2001database}, Urban100\cite{huang2015single} and Manga109\cite{matsui2017sketch}. The restoration quality of the resulted SR is evaluated by using PSNR and SSIM\cite{wang2004image}.

  Three scaling cases \{ $ 2\times, 4\times, 8\times$ \} are considered. On each case, the architecture of feature extraction part of our network is kept the same, and the transposed convolutional layer size is changed according to different up-scale rate. The source code of our method is available on GitHub\footnote{source code: \url{https://github.com/MKFMIKU/SrSENet}}.

\subsection{Training Details}
\begin{figure}
    \setlength{\abovecaptionskip}{0.cm}
    \setlength{\belowcaptionskip}{-0.cm}
    \begin{center}
    \begin{tabular}{@{}ccc}
    \includegraphics[width=0.33\textwidth]{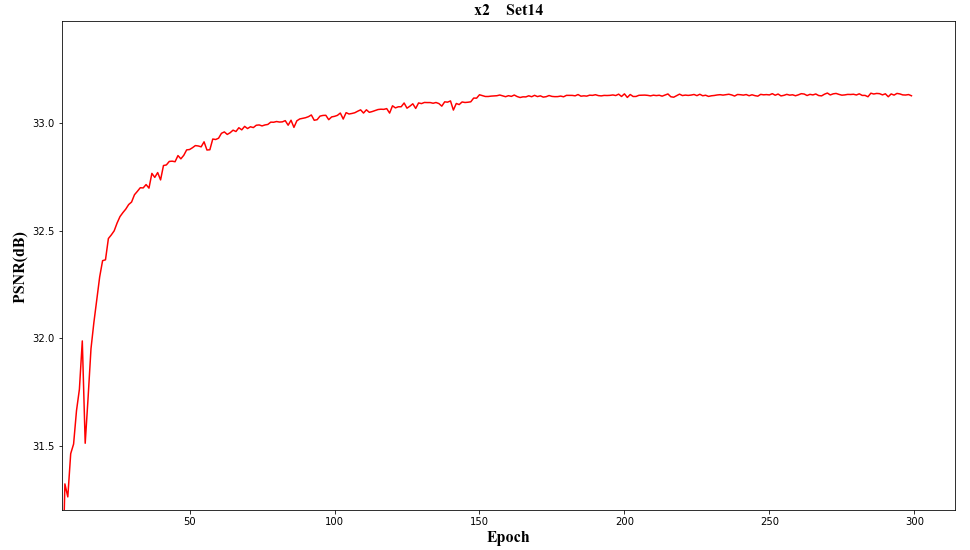}
    &\includegraphics[width=0.33\textwidth]{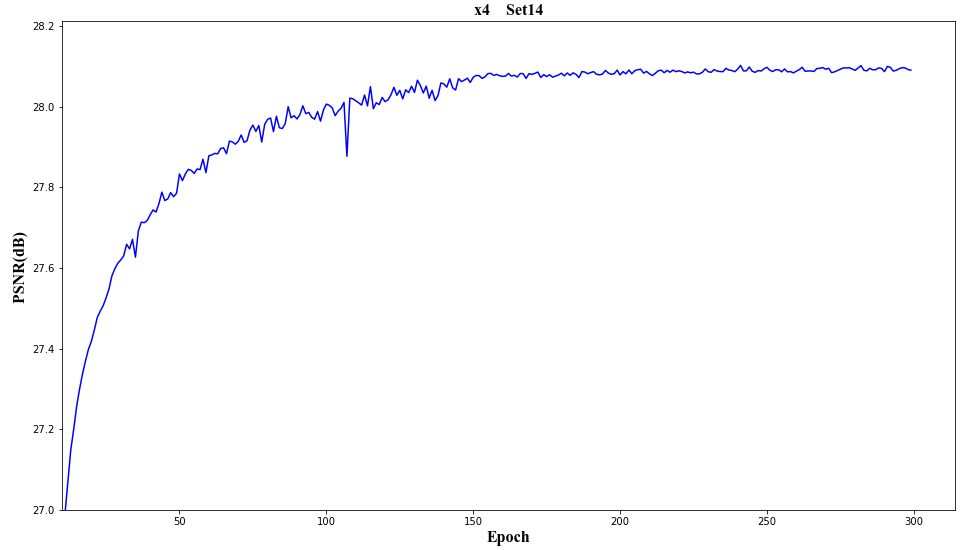}
    &\includegraphics[width=0.33\textwidth]{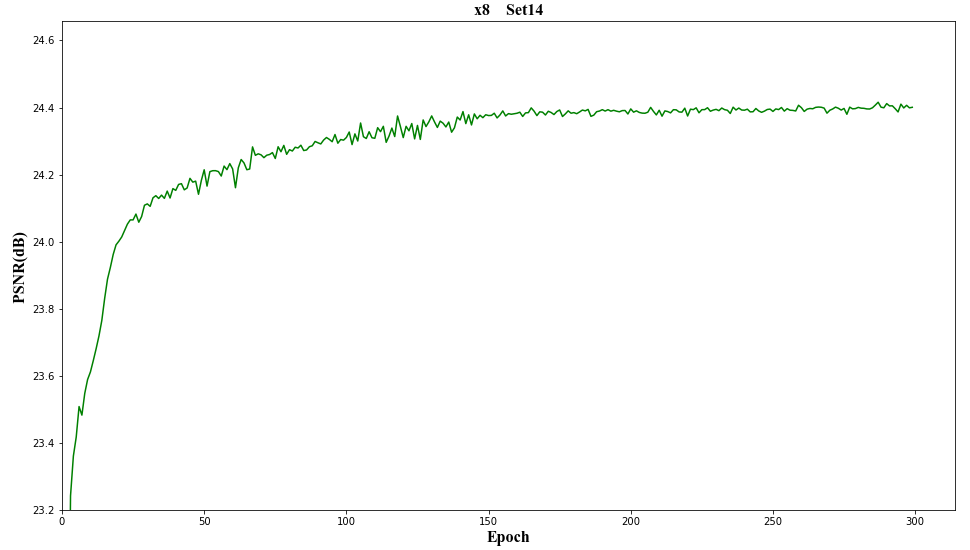}
    \end{tabular}
    \end{center}
    \caption{Respective PNSR testing curves of SrSENet on dataset Set14 for three scaling cases. Left: scale $2\times$, Middle: scale $4\times$, Right: scale $8\times$.}
    \label{testcurve}
\end{figure}

  We use 8 SrSEBlocks to do feature extraction. For each upscale deconvolution layer, we use respective convolutional kernels [4,2,1], [8,4,2], [16,8,4] for 2x, 4x, 8x rate up-scaled super-resolution image respectively. Here in the format $[*,*,*]$, the first represents kernel size, the second represents stride steps, and the last is padding size in transposed layer. If dealing with odd multiples of magnification, we can also easily achieve an odd magnification by modifying the kernel size of the convolutional network to an odd number(e.g., [3,*,*]). During the training, we set the initial learning rate at $0.0001$. We use Adma optimizer~\cite{kingma2014adam} with $\beta_{1} = 0.9$ to let network convergence and the training batches is 64. It roughly takes half day on a machine using four TitanX GPUs for a single upscale training. For illustration, the respective PSNR testing curves of our SrSENet on Set14 are shown in Figure~\ref{testcurve}.

  The quantitative performance comparisons are shown in Table~\ref{SRResult}. From the experiment results, we can easily find that our proposed method obtains competitive performance in all datasets in different upscale rates. Especially in larger scale case, the advantages of our method are more obvious. Our method can achieve top performance with less network depth.
  In Figure~\ref{Visual}, we further show some realistic results for visual comparison. We can find that the fine texture of images in our method are recovered better.

\begin{table*}[h]
  \small
  \setlength{\abovecaptionskip}{0.cm}
  \setlength{\belowcaptionskip}{-0.cm}
  \caption{Quantitative comparisons of state-of-the-art methods. \textcolor[rgb]{1.00,0.00,0.00}{Red} text indicates the best performance and \textcolor[rgb]{0.00,0.00,1.00}{blue} italics text indicates the second best performance. We use results from LapSRN to do comparation, and attention that Layers in the table include convolution and deconvolution.}
  \label{SRResult}
  \begin{center}
  \begin{tabular}{lccccccc}
  \hline
  \multicolumn{1}{c}{Algorithm} & Scale & \begin{tabular}[c]{@{}c@{}}Set5\\PSNR/SSIM\end{tabular} & \begin{tabular}[c]{@{}c@{}}Set14\\PSNR/SSIM\end{tabular} & \begin{tabular}[c]{@{}c@{}}BSDS100\\PSNR/SSIM\end{tabular} & \begin{tabular}[c]{@{}c@{}}Urban100\\PSNR/SSIM\end{tabular} & \begin{tabular}[c]{@{}c@{}}Manga109\\PSNR/SSIM\end{tabular}\\ \hline
  Bicubic & 2x & 33.65/0.930 & 30.34/0.870 & 29.56/0.844 & 26.88/0.841 & 30.84/0.935 \\
  SelfExSR~\cite{huang2015single} & 2x  & 36.49/0.954 & 32.44/0.906 & 31.18/0.886 & 29.54/0.897 & 35.78/0.968 \\
  SRCNN~\cite{dong2014learning}   & 2x  & 36.65/0.954 & 32.29/0.903 & 31.36/0.888 & 29.52/0.895 & 35.72/0.968 \\
  FSRCNN~\cite{dong2016accelerating}   & 2x & 36.99/0.955 & 32.73/0.909 & 31.51/0.891 & 29.87/0.901 & 36.62/0.971 \\
  VDSR~\cite{kim2016accurate}   & 2x & 37.53/\emph{\color{blue} 0.958} & 32.97/\color{red}0.913  & \color{red} 31.90/\color{red} 0.896 & \color{red} 30.77/\emph{\color{blue} 0.914} & 37.16/\color{red} 0.974 \\
  DRCN~\cite{kim2016deeply}  & 2x & \color{red} 37.63/\color{red} 0.959 & 32.98/\color{red}0.913 &  \emph{\color{blue} 31.85}/0.894 &  \emph{\color{blue} 30.76}/0.913 &  \color{red} 37.57/\emph{\color{blue}0.973} \\
  LapSRN~\cite{LapSRN}  & 2x & 37.52/\color{red} 0.959 & \emph{\color{blue} 33.08}/\color{red} 0.913 & 31.80/\emph{\color{blue}0.895} & 30.41/0.910 &  37.27/\color{red}0.974 \\
  SrSENet & 2x & \emph{\color{blue} 37.56}/\emph{\color{blue} 0.958} &  \color{red} 33.14/\emph{\color{blue} 0.911} & 31.84/\color{red} 0.896 &   30.73/\color{red} 0.917 & \emph{\color{blue} 37.43}/\color{red} 0.974 \\ \hline

  Bicubic & 4x & 28.42/0.810 & 26.10/0.704 & 25.96/0.669 & 23.15/0.659 & 24.92/0.789 \\
  SelfExSR~\cite{huang2015single} & 4x & 30.33/0.861 & 27.54/0.756 & 26.84/0.712 & 24.82/0.740 & 27.82/0.865 \\
  SRCNN~\cite{dong2014learning}   & 4x & 30.49/0.862 & 27.61/0.754 & 26.91/0.712 & 24.53/0.724 & 27.66/0.858 \\
  FSRCNN~\cite{dong2016accelerating}   & 4x & 30.71/0.865 & 27.70/0.756 & 26.97/0.714 & 24.61/0.727 & 27.89/0.859 \\

  VDSR~\cite{kim2016accurate}    & 4x & 31.35/0.882 & 28.03/\emph{\color{blue}0.770}  &  \emph{\color{blue} 27.29}/\emph{\color{blue} 0.726} &  \emph{\color{blue} 25.18}/0.753 & 28.82/0.886 \\
  DRCN~\cite{kim2016accurate}    & 4x &  \emph{\color{blue} 31.53}/\emph{\color{blue}0.884} & 28.04/\emph{\color{blue}0.770} & 27.24/0.724 &  25.14/0.752 & 28.97/0.886 \\
  LapSRN~\cite{LapSRN}  & 4x &  \color{red} 31.54/\color{red} 0.885 &   \color{red} 28.19/\color{red} 0.772 & \color{red} 27.32/\color{red}0.728 &   \color{red} 25.21/\emph{\color{blue} 0.756} &  \color{red} 29.09/\color{red} 0.890 \\
  SrSENet & 4x & 31.40/0.881 &  \emph{\color{blue}28.10}/0.766 & \emph{\color{blue} 27.29}/0.720 &  \color{red} 25.21/\color{red} 0.762 & \emph{\color{blue} 29.08}/\emph{\color{blue}0.888} \\ \hline

  Bicubic & 8x & 24.40/0.657 & 23.19/0.568 & 23.67/0.547 & 20.74/0.515 & 21.47/0.649 \\
  SelfExSR~\cite{huang2015single} & 8x & 25.52/0.704 & 24.02/0.603 & 24.18/0.568 &  \emph{\color{blue}21.81}/\emph{\color{blue}0.576} & 22.99/0.718  \\
  SRCNN~\cite{dong2014learning}   & 8x & 25.33/0.689 & 23.85/0.593 & 24.13/0.565 & 21.29/0.543 & 22.37/0.682 \\
  FSRCNN~\cite{dong2016accelerating}      & 8x & 25.41/0.682 & 23.93/0.592 & 24.21/0.567 & 21.32/0.537 & 22.39/0.672 \\
  VDSR~\cite{kim2016accurate}    & 8x  & 25.72/\emph{\color{blue}0.711} & 24.21/\emph{\color{blue}0.609} & 24.37/\emph{\color{blue}0.576} & 21.54/0.560 & 22.83/0.707 \\
  LapSRN~\cite{LapSRN}  & 8x & \color{red} 26.14/\color{red} 0.738 &   \color{red} 24.44/\emph{\color{blue} 0.623} &  \emph{\color{blue} 24.54}/\color{red} 0.586 &   \emph{\color{blue} 21.81}/\emph{\color{blue} 0.581} &   \emph{\color{blue} 23.39}/\color{red} 0.735 \\
  SrSENet & 8x & \emph{\color{blue} 26.10}/0.703 & \emph{\color{blue} 24.38}/0.586 & \color{red} 24.59/0.539 &  \color{red} 21.88/0.571 & \color{red} 23.54/\emph{\color{blue} 0.722} \\\hline
  \end{tabular}
  \end{center}
  \end{table*}

  \begin{figure}
  \setlength{\abovecaptionskip}{0.cm}
  \setlength{\belowcaptionskip}{-0.cm}
  \begin{center}
  \small
  \begin{tabular}{@{}lccccccc}
  \multirow{15}{*}{\begin{tabular}[c]{@{}c@{}}\includegraphics[width=0.11\textwidth]{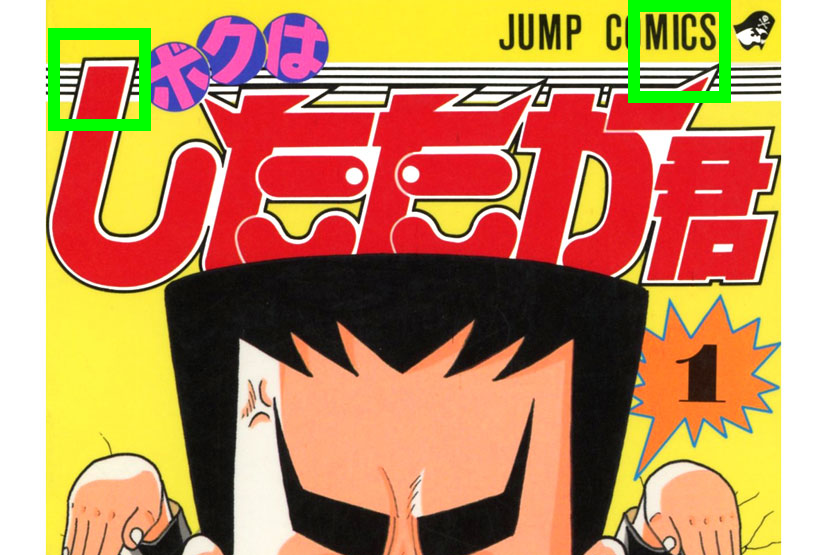}\\HR x8\\Manga109\end{tabular}} \\ \\ \\
  &\includegraphics[width=0.12\textwidth]{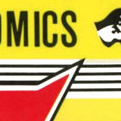}
  &\includegraphics[width=0.12\textwidth]{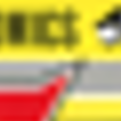}
  &\includegraphics[width=0.12\textwidth]{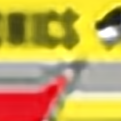}
  &\includegraphics[width=0.12\textwidth]{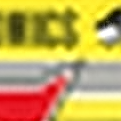}
  &\includegraphics[width=0.12\textwidth]{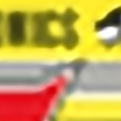}
  &\includegraphics[width=0.12\textwidth]{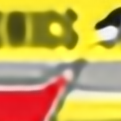}
  &\includegraphics[width=0.12\textwidth]{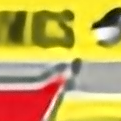}
  \\
  & GT & Bicubic & SRCNN  & FSRCNN & VDSR & LapSRN & SrSENet
  \\
  &\includegraphics[width=0.12\textwidth]{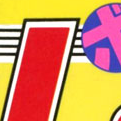}
  &\includegraphics[width=0.12\textwidth]{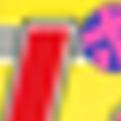}
  &\includegraphics[width=0.12\textwidth]{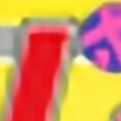}
  &\includegraphics[width=0.12\textwidth]{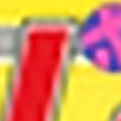}
  &\includegraphics[width=0.12\textwidth]{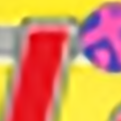}
  &\includegraphics[width=0.12\textwidth]{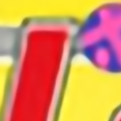}
  &\includegraphics[width=0.12\textwidth]{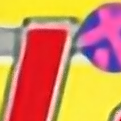}
  \\
  & GT & Bicubic & SRCNN  & FSRCNN & VDSR & LapSRN & SrSENet
  \\
  \end{tabular}

  \begin{tabular}{@{}lccccccc}
  \multirow{15}{*}{\begin{tabular}[c]{@{}c@{}}\includegraphics[width=0.12\textwidth]{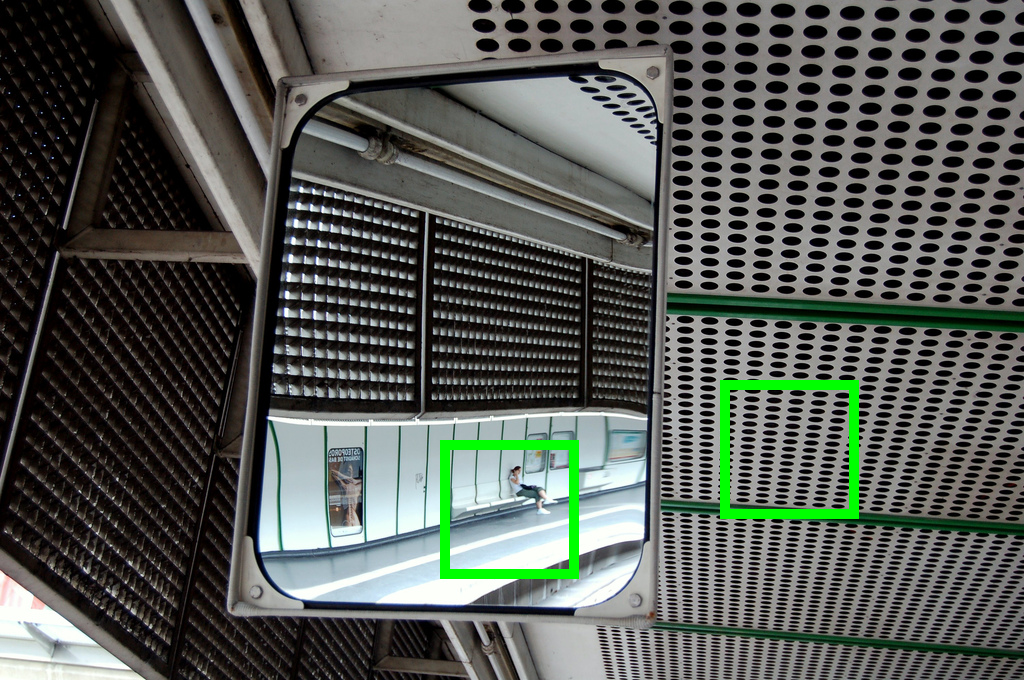}\\HR x4\\Urban100\end{tabular}}\\ \\ \\
  &\includegraphics[width=0.12\textwidth]{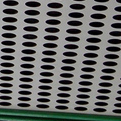}
  &\includegraphics[width=0.12\textwidth]{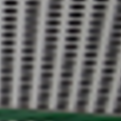}
  &\includegraphics[width=0.12\textwidth]{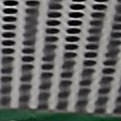}
  &\includegraphics[width=0.12\textwidth]{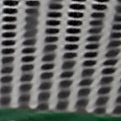}
  &\includegraphics[width=0.12\textwidth]{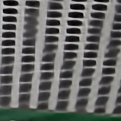}
  &\includegraphics[width=0.12\textwidth]{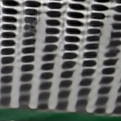}
  &\includegraphics[width=0.12\textwidth]{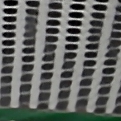}
  \\
  & GT & Bicubic & SRCNN  & FSRCNN & VDSR & LapSRN & SrSENet
  \\
  &\includegraphics[width=0.12\textwidth]{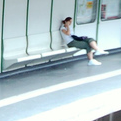}
  &\includegraphics[width=0.12\textwidth]{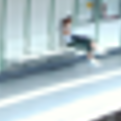}
  &\includegraphics[width=0.12\textwidth]{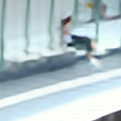}
  &\includegraphics[width=0.12\textwidth]{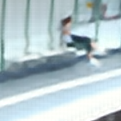}
  &\includegraphics[width=0.12\textwidth]{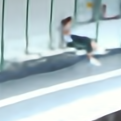}
  &\includegraphics[width=0.12\textwidth]{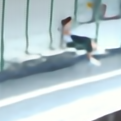}
  &\includegraphics[width=0.12\textwidth]{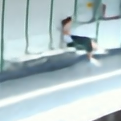}
  \\
  & GT & Bicubic & SRCNN  & FSRCNN & VDSR & LapSRN & SrSENet
  \\
  \end{tabular}

  \begin{tabular}{@{}lccccccc}
  \multirow{15}{*}{\begin{tabular}[c]{@{}c@{}}\includegraphics[width=0.12\textwidth]{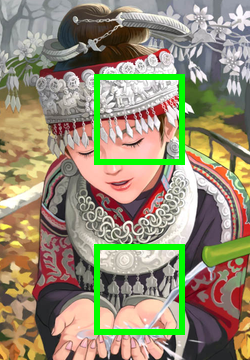}\\HR x2\\Set14\end{tabular}}\\ \\ \\
  &\includegraphics[width=0.12\textwidth]{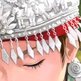}
  &\includegraphics[width=0.12\textwidth]{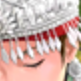}
  &\includegraphics[width=0.12\textwidth]{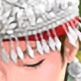}
  &\includegraphics[width=0.12\textwidth]{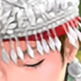}
  &\includegraphics[width=0.12\textwidth]{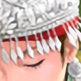}
  &\includegraphics[width=0.12\textwidth]{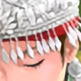}
  &\includegraphics[width=0.12\textwidth]{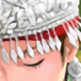}
  \\
  & GT & Bicubic & SRCNN  & FSRCNN & VDSR & LapSRN & SrSENet
  \\
  &\includegraphics[width=0.12\textwidth]{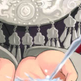}
  &\includegraphics[width=0.12\textwidth]{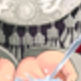}
  &\includegraphics[width=0.12\textwidth]{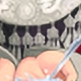}
  &\includegraphics[width=0.12\textwidth]{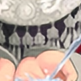}
  &\includegraphics[width=0.12\textwidth]{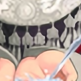}
  &\includegraphics[width=0.12\textwidth]{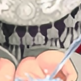}
  &\includegraphics[width=0.12\textwidth]{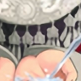}
  \\
  & GT & Bicubic & SRCNN  & FSRCNN & VDSR & LapSRN & SrSENet
  \\
  \end{tabular}
  \end{center}
  \caption{ Visual comparisons on Bicubic, SRCNN, FSRCNN, VDSR, LapSRN and SrSENet on upscale rate of $8\times$, $4\times$, $2\times$.}
  \label{Visual}
  \end{figure}

  \begin{table*}[!h]
    \small
    \setlength{\abovecaptionskip}{0.cm}
    \setlength{\belowcaptionskip}{-0.cm}
    \caption{Ablation experiment: quantitative comparisons on $4 \times$ scale.}
    \label{ablation}
    \begin{center}
    \begin{tabular}{lcccccc}
    \hline
    \multicolumn{1}{c}{Algorithm} & \begin{tabular}[c]{@{}c@{}}Set5\\PSNR/SSIM\end{tabular} & \begin{tabular}[c]{@{}c@{}}Set14\\PSNR/SSIM\end{tabular} & \begin{tabular}[c]{@{}c@{}}BSDS100\\PSNR/SSIM\end{tabular} & \begin{tabular}[c]{@{}c@{}}Urban100\\PSNR/SSIM\end{tabular} & \begin{tabular}[c]{@{}c@{}}Manga109\\PSNR/SSIM\end{tabular}\\ \hline
    reduced version & 31.30/0.880 & 28.10/0.766 & 27.16/0.720 & 25.08/0.760 & 28.84/0.886\\
    SrSENet & 31.40/0.881 &  28.10/0.766 & 27.29/0.720 & 25.21/0.762 & 29.08/0.888 \\
    \hline
    \end{tabular}
    \end{center}
    \end{table*}

  In order to verify the effectiveness of the introduced SEblock, we additionally have set up an ablation experiment. We construct a reduced version network by removing SElayers out of the proposed SrSENet, while keeping other parts remained. We have compared the reduced version with our SrSENet. The performance comparisons on $4 \times$ scale are shown in Table~\ref{ablation}. From the results, we can easily find that the introduced SEblocks indeed plays great importance on final excellence performance. On the other scales, we could achieve similar conclusions as well. We owe the its effectiveness to it introducing channel-wise attention mechanism, which makes channel information of each pixel on SR image adaptively learnable.

  \section{Conclusions}
  In this paper, we have proposed a new effective super-resolution model by using a deep residual network with SrSEBlock. Our method focuses on modeling channels correlations between feature mappings from the LR image. By modeling channel wise, we have confirmed that our method could produce more realistic texture on realistic images. We set a new state-of-the-art super-resolution method without increasing the complexities of the network. We believe that our approach can be applied to other real-world computer vision problems and achieve competitive results.

  \section*{Acknowledgment}
  This work was supported by National Natural Science Foundation of China under Grant Nos. 61365002, 61462042 and 61462045.


\begin{thebibliography}{99}
\providecommand{\url}[1]{\texttt{#1}}
\providecommand{\urlprefix}{URL }
\providecommand{\doi}[1]{https://doi.org/#1}

\bibitem{Agustsson_2017_CVPR_Workshops}
Agustsson, E., Timofte, R.: {NTIRE 2017 Challenge on Single Image
  Super-Resolution: Dataset and Study}. In: Proceedings of the IEEE Conference
  on Computer Vision and Pattern Recognition Workshops. pp. 1110--1121. IEEE,
  Hawaii (2017)

\bibitem{bruhn2005lucas}
Bruhn, A., Weickert, J., Schn{\"{o}}rr, C.: {Lucas/Kanade meets Horn/Schunck:
  Combining local and global optic flow methods}. International Journal of
  Computer Vision  \textbf{61}(3),  211--231 (2005)

\bibitem{dong2014learning}
Dong, C., Loy, C.C., He, K., Tang, X.: {Learning a deep convolutional network
  for image super-resolution}. In: Proceedings of the European Conference on
  Computer Vision. pp. 184--199. Springer, Zurich (2014)

\bibitem{dong2016accelerating}
Dong, C., Loy, C.C., Tang, X.: {Accelerating the super-resolution convolutional
  neural network}. In: Proceedings of the European Conference on Computer
  Vision. pp. 391--407. Springer, Amsterdam (2016)

\bibitem{he2016deep}
He, K., Zhang, X., Ren, S., Sun, J.: {Deep residual learning for image
  recognition}. In: Proceedings of the IEEE Conference on Computer Vision and
  Pattern Recognition. pp. 770--778. IEEE, Las Vegas (2016)

\bibitem{hu2017squeeze}
Hu, J., Shen, L., Sun, G.: {Squeeze-and-excitation networks}. In: Proceedings
  of the IEEE Conference on Computer Vision and Pattern Recognition. IEEE, Salt
  Lake (2018)

\bibitem{huang2015single}
Huang, J.B., Singh, A., Ahuja, N.: {Single image super-resolution from
  transformed self-exemplars}. In: Proceedings of the IEEE Conference on
  Computer Vision and Pattern Recognition. pp. 5197--5206. IEEE, Boston (2015)

\bibitem{kim2016accurate}
Kim, J., {Kwon Lee}, J., {Mu Lee}, K.: {Accurate image super-resolution using
  very deep convolutional networks}. In: Proceedings of the IEEE Conference on
  Computer Vision and Pattern Recognition. pp. 1646--1654. IEEE, Las Vegas
  (2016)

\bibitem{kim2016deeply}
Kim, J., {Kwon Lee}, J., {Mu Lee}, K.: {Deeply-recursive convolutional network
  for image super-resolution}. In: Proceedings of the IEEE Conference on
  Computer Vision and Pattern Recognition. pp. 1637--1645. IEEE, Las Vegas
  (2016)

\bibitem{kingma2014adam}
Kingma, D., Ba, J.: {Adam: A method for stochastic optimization}. In:
  Proceedings of the International Conference on Learning Representations. San
  Diego (2015)

\bibitem{LapSRN}
Lai, W.S., Huang, J.B., Ahuja, N., Yang, M.H.: {Deep Laplacian Pyramid Networks
  for Fast and Accurate Super-Resolution}. In: Proceedings of the IEEE
  Conference on Computer Vision and Pattern Recognition. pp. 5835--5843. IEEE,
  Hawaii (2017)

\bibitem{ledig2016photo}
Ledig, C., Theis, L., Husz{\'{a}}r, F., Caballero, J., Cunningham, A., Acosta,
  A., Aitken, A., Tejani, A., Totz, J., Wang, Z., Others: {Photo-realistic
  single image super-resolution using a generative adversarial network}. In:
  Proceedings of the IEEE Conference on Computer Vision and Pattern
  Recognition. pp. 105--114. IEEE, Hawaii (2017)

\bibitem{martin2001database}
Martin, D., Fowlkes, C., Tal, D., Malik, J.: {A database of human segmented
  natural images and its application to evaluating segmentation algorithms and
  measuring ecological statistics}. In: Proceedings of the IEEE International
  Conference on Computer Vision. pp. 416--423. IEEE, Vancouver (2001)

\bibitem{matsui2017sketch}
Matsui, Y., Ito, K., Aramaki, Y., Fujimoto, A., Ogawa, T., Yamasaki, T.,
  Aizawa, K.: {Sketch-based manga retrieval using manga109 dataset}. Multimedia
  Tools and Applications  \textbf{76}(20),  21811--21838 (2017)

\bibitem{schulter2015fast}
Schulter, S., Leistner, C., Bischof, H.: {Fast and accurate image upscaling
  with super-resolution forests}. In: Proceedings of the IEEE Conference on
  Computer Vision and Pattern Recognition. pp. 3791--3799. IEEE, Boston (2015)

\bibitem{shi2016real}
Shi, W., Caballero, J., Husz{\'{a}}r, F., Totz, J., Aitken, A.P., Bishop, R.,
  Rueckert, D., Wang, Z.: {Real-time single image and video super-resolution
  using an efficient sub-pixel convolutional neural network}. In: Proceedings
  of the IEEE Conference on Computer Vision and Pattern Recognition. pp.
  1874--1883. IEEE, Las Vegas (2016)

\bibitem{timofte2014a+}
Timofte, R., {De Smet}, V., {Van Gool}, L.: {A+: Adjusted anchored neighborhood
  regression for fast super-resolution}. In: Proceedings of the 12th Asian
  Conference on Computer Vision. pp. 111--126. Springer, Singapore (2014)

\bibitem{wang2004image}
Wang, Z., Bovik, A.C., Sheikh, H.R., Simoncelli, E.P.: {Image quality
  assessment: from error visibility to structural similarity}. IEEE
  Transactions on Image Processing  \textbf{13}(4),  600--612 (2004)

\bibitem{Yang2013}
Yang, C.Y., Yang, M.H.: {Fast Direct Super-Resolution by Simple Functions}. In:
  Proceedings of the IEEE International Conference on Computer Vision. pp.
  561--568. IEEE, Sydney (2013)

\bibitem{yang2008image}
Yang, J., Wright, J., Huang, T., Ma, Y.: {Image super-resolution as sparse
  representation of raw image patches}. In: Proceedings of the IEEE Conference
  on Computer Vision and Pattern Recognition. pp.~1--8. IEEE, Anchorage (2008)

\bibitem{yang2010image}
Yang, J., Wright, J., Huang, T.S., Ma, Y.: {Image super-resolution via sparse
  representation}. IEEE transactions on image processing  \textbf{19}(11),
  2861--2873 (2010)

\bibitem{zeyde2010single}
Zeyde, R., Elad, M., Protter, M.: {On single image scale-up using
  sparse-representations}. In: Proceedings of the International Conference on
  Curves and Surfaces. pp. 711--730. Springer, Avignon (2010)

\end{thebibliography}

\end{document}